\documentclass[journal]{IEEEtran}

\usepackage{cite}
\usepackage{amsmath,amssymb}
\usepackage{newtxtext,newtxmath}
\usepackage{graphicx}
\usepackage{booktabs}
\usepackage{array}
\usepackage{algorithm}
\usepackage{algpseudocode}
\usepackage{url}
\usepackage{microtype}
\usepackage{balance}

\newcommand{\rT}{r_T}
\DeclareMathOperator{\clip}{clip}
\DeclareMathOperator{\Var}{Var}

\begin{document}

\title{TriNetra: Thermal-Reliability-Guided RGB-Thermal\\Fusion and Tracking for Surveillance}

\newcommand{\authorcell}[2]{\begin{minipage}[t]{0.31\textwidth}\centering
#1\\[1pt]{\footnotesize Dept. of Computer Science and Design\\
K. K. Wagh Institute of Engineering\\
Education and Research\\
Nashik 422003, India\\
#2}\end{minipage}}

\author{%
\authorcell{Dr. Y. D. Bhise}{ydbhise@kkwagh.edu.in}\hfill
\authorcell{Aditya Ahirrao}{adityaahirrao17@gmail.com}\hfill
\authorcell{Disha Kapse}{dishakapse55@gmail.com}\\[8pt]
\authorcell{Sadique Khatib}{khatibsadik9@gmail.com}\hspace{0.08\textwidth}
\authorcell{Vedant Sonawane}{vedantsonawane2244@gmail.com}}

\markboth{TRINETRA: THERMAL-RELIABILITY-GUIDED FUSION AND TRACKING}{}

\maketitle

\begin{abstract}
Visible cameras fail at night and thermal sensors fail under saturation, blur or loss of contrast, yet most RGB-thermal (RGB-T) detectors mix the two streams with fixed weights and cannot tell when one modality has become unreliable. TriNetra is a reliability-aware surveillance framework built around this problem. The proposed solution computes a frame-level Thermal Reliability Coefficient (TRC) from five label-free image cues, uses it to gate the thermal features of a dual-branch YOLOv8m detector whose visible branch, neck and head are frozen and COCO-pretrained, and passes the same score to a ByteTrack-based tracker that feeds a rule-based behaviour analyser. On the M3FD benchmark (2,940/630/630 split) the detector reaches 87.6\% precision, 79.4\% recall, 70.0\% mAP@50 and 47.8\% mAP@50:95 on the test split, with 17.0\,M of its 42.9\,M parameters trained. TRC stays near 0.78 on clean frames (standard deviation 0.012), decreases with the severity of thermal saturation and blur, reaching 0.53 and 0.69 at full severity, and is insensitive to RGB-thermal misalignment. With identical weights, the adaptive gate matches a fixed gate on clean data (70.0\% versus 69.8\% mAP@50) but keeps 43.6\% mAP@50 when the thermal stream is corrupted, against 38.4\%, 29.2\% and 25.0\% for a fixed gate, full thermal admission and a closed gate; the 5.1-point gain over the fixed gate has a 95\% bootstrap confidence interval of $+3.6$ to $+6.5$ points. In tracking, the TRC-gated association performs on par with ByteTrack on a synthetic benchmark with exact ground truth and on 14 M3FD clips, and is slightly less stable under thermal saturation, so its measured benefit is confined to the detector. The limitations of the current evaluation are reported together with the results.
\end{abstract}


\section{Introduction}
\IEEEPARstart{D}{etecting} and following people and vehicles around the clock is hard for a visible-light camera, which loses contrast in darkness, glare, fog, smoke and rain. Thermal infrared cameras record emitted rather than reflected radiation and remain informative in these conditions, which has made paired RGB-thermal (RGB-T) sensing the usual route to all-weather perception \cite{songSurvey,deepFusion}.

Thermal imagery is not uniformly reliable either. Sensor saturation, defocus, motion blur, loss of contrast at thermal crossover and RGB-thermal misregistration all reduce its value, and a detector that mixes the two streams with fixed weights inherits the damage \cite{denet,qdfm}. Recent RGB-T detectors improve the fusion itself through target-aware feature refinement \cite{tfdet}, adaptive gated fusion in a lightweight detector \cite{gemyolo} and energy-score-guided enhancement \cite{eef}. Other work addresses failures directly, through diffusion-based reconstruction with dynamic modality weighting \cite{denet} or through quality-aware decision fusion based on Dempster--Shafer theory \cite{qdfm}. In these methods reliability is learned implicitly and stays inside the network weights, so it cannot be inspected, cannot be checked against a known degradation, and is not passed on to later stages.

Two gaps follow. First, none of the works above exposes an explicit, label-free, per-frame estimate of thermal reliability that can be measured against controlled degradations. Second, RGB-T detection is usually evaluated in isolation, whereas surveillance needs persistent identities and behaviour cues built on top of the detections.

This article presents TriNetra, a reliability-aware framework for RGB-T surveillance. The proposed solution conditions each pair of frames, computes a Thermal Reliability Coefficient (TRC) from five image cues, uses the TRC to scale the thermal features of a dual-branch YOLOv8 detector \cite{yolov8} whose visible branch, neck and head stay frozen, and passes the same score to a ByteTrack-style tracker \cite{bytetrack} that feeds a rule-based behaviour analyser. The contributions are as follows.
\begin{enumerate}
\item \textit{A label-free reliability score.} The TRC combines contrast, sharpness, saturation, RGB-thermal mutual information and alignment quality into one number in $[0,1]$. It is characterised on all 630 M3FD test pairs and on four controlled degradation types at four severities, including a leave-one-cue-out analysis.
\item \textit{A TRC-gated detector.} The dual-branch detector trains 17.02\,M of 42.92\,M parameters and reaches 70.0\% mAP@50 and 47.8\% mAP@50:95 on the M3FD test split.
\item \textit{A gate ablation under corrupted thermal input.} With identical weights, the adaptive gate equals a fixed gate on clean data but retains 43.6\% mAP@50 under thermal corruption, against 38.4\%, 29.2\% and 25.0\% for a fixed gate, full thermal admission and a closed gate. Bootstrap confidence intervals over the test images are reported for every comparison.
\item \textit{Reliability-aware tracking and behaviour analysis with a four-layer validation protocol.} The layers are unit tests, a synthetic benchmark with exact ground truth, label-free metrics on the clips and a thermal-degradation stress test; in this evaluation the TRC-gated association performs on par with ByteTrack and slightly worse under thermal saturation.
\end{enumerate}

The remainder of this article is organised as follows. Section~II reviews related work. Section~III presents the proposed solution. Section~IV describes the datasets, environment, metrics and protocols. Section~V reports and discusses the results, and Section~VI concludes.

\section{Related Work}

\subsection{RGB-T Object Detection and Multimodal Fusion}
RGB-T analysis exploits complementary visible and thermal measurements and has been studied for detection, segmentation, fusion, tracking and surveillance \cite{songSurvey}. Multimodal fusion in general is surveyed in \cite{deepFusion}. Single-stage detectors of the YOLO family \cite{yolov8,yolo} are a common backbone for RGB-T detection because of their speed, and TriNetra also builds on YOLOv8m.

TFDet proposes target-aware fusion for RGB-T pedestrian detection: a fusion-refinement design examines parallel and cross-channel similarities, and a segmentation branch separates target-related features from background features \cite{tfdet}. GEM-YOLO combines an adaptive multimodal gated fusion mechanism with lightweight components for real-time RGB-T detection \cite{gemyolo}. Fixed or insufficiently adaptive fusion is a shared concern with TriNetra, but in GEM-YOLO the gate is learned inside the network, whereas TriNetra drives its gate with an explicit, hand-designed and inspectable reliability score. EEF introduces energy-score-guided feature enhancement, efficient cross-modal fusion and adaptive feedback \cite{eef}. TriNetra instead uses a low-cost image-statistics signal that can be examined independently of detector confidence.

\subsection{Robustness to Modality Failure and Quality-Aware Fusion}
Wang \textit{et al.} study RGB-T detection under failure scenarios and propose DENet, which combines diffusion-based information reconstruction, redundant-information suppression and a kernel-similarity-aware illumination module for dynamic modality weighting \cite{denet}. Yu \textit{et al.} propose quality-aware decision fusion based on evidential deep learning and Dempster--Shafer theory, which expresses uncertainty at the decision level and improves accuracy under sensor interference and failure \cite{qdfm}. TriNetra shares the quality-awareness objective but acts earlier: it estimates thermal reliability from the raw thermal statistics before feature fusion, needs no labelled failure examples, and exposes a scalar that can be plotted against a known degradation. The price is that a hand-designed score only measures the failure types its cues cover.

\subsection{Multi-Object Tracking for Surveillance}
SORT associates detections with Kalman-predicted tracks through IoU and the Hungarian algorithm \cite{sort,kalman,kuhn}. DeepSORT adds an appearance metric \cite{deepsort}. ByteTrack observes that low-confidence detections often belong to partly occluded objects and associates them in a second stage instead of discarding them \cite{bytetrack}. Tracking quality is commonly scored with HOTA, CLEAR MOT and IDF1 \cite{hota,clearmot,idf1}. In the RGB-T detection works reviewed above, no reliability signal is passed on to a tracker. TriNetra tests one simple mechanism: when the frame's TRC is low, the association cost of low-confidence boxes is inflated so that such boxes must agree more closely with the motion prediction.

\subsection{Positioning}
TriNetra is not a new fusion backbone. Its contribution is the explicit reliability signal, its use in both the detector and the tracker, and an evaluation that separates what the signal achieves (detection under thermal corruption) from what it does not (tracking accuracy in the present tests).

\section{Proposed Solution}

\subsection{Overview}
\begin{figure*}[!t]
\centering
\includegraphics[width=\textwidth]{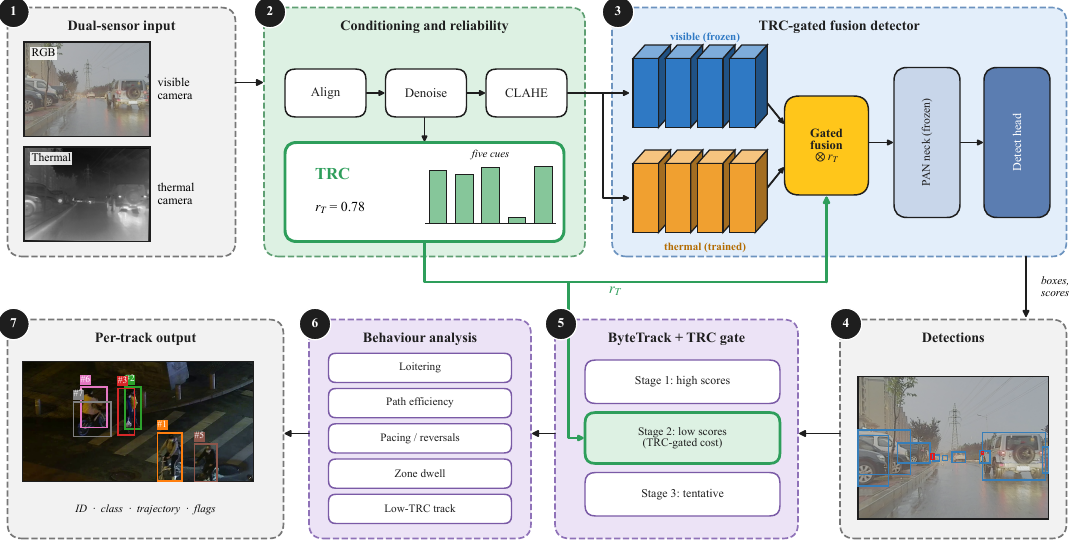}
\caption{Overall TriNetra pipeline. The reliability score $\rT$ (green) drives both the gated fusion inside the detector and the association gate inside the tracker. Numbers give the processing order; the example frames are an M3FD test pair (detections) and a night clip (tracks).}
\label{fig:pipeline}
\end{figure*}

Fig.~\ref{fig:pipeline} shows the pipeline. Paired RGB and thermal frames are conditioned (alignment, denoising, normalisation), the TRC is computed on the conditioned thermal frame, and the dual-branch detector fuses the two streams under control of the TRC. Detections and the TRC enter the tracker, whose confirmed tracks feed the behaviour analyser. A threat-reasoning stage is planned but is not part of the present evaluation.

\subsection{Preprocessing and Alignment}
The visible frame is used as the spatial reference. Thermal alignment is the identity when the pair is pre-registered (as in M3FD \cite{m3fd}) and a homography otherwise; the alignment step also reports a reprojection error $d$ in pixels (zero for the identity). Both frames are denoised with a bilateral filter (diameter 5, $\sigma=10$) \cite{bilateral}. Intensity is then normalised with contrast-limited adaptive histogram equalisation (clip limit 2.0, $8\times 8$ tiles) \cite{clahe}, and the thermal frame is first clipped to its 1st--99th percentile range to suppress outliers. The conditioned frames feed both the TRC and the detector.

\subsection{Thermal Reliability Coefficient}
\begin{figure*}[!t]
\centering
\includegraphics[width=\textwidth]{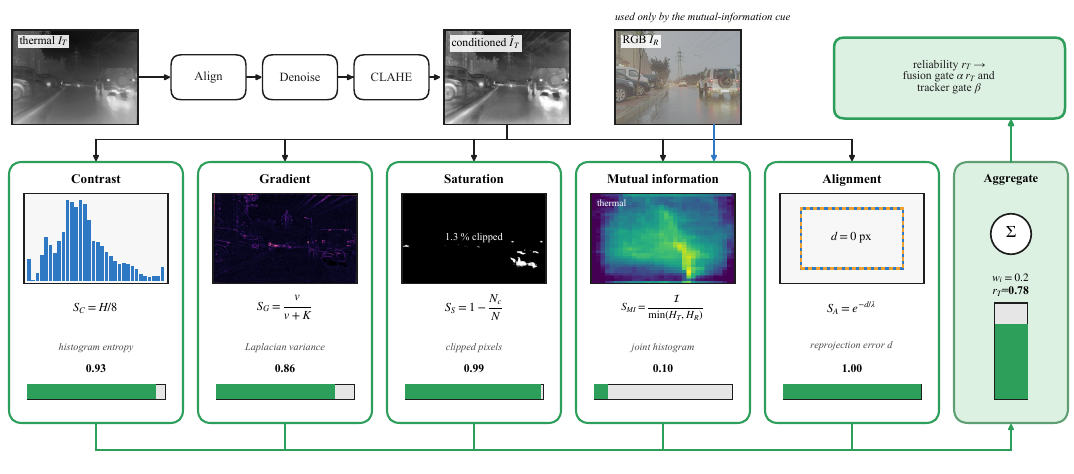}
\caption{Computation of the Thermal Reliability Coefficient. The thermal frame passes through alignment, denoising and normalisation; five cues are computed per frame and aggregated into $\rT$. The cue maps (histogram, Laplacian, clipped pixels, joint histogram) and scores are computed from one M3FD test pair. The RGB frame is needed only by the mutual-information cue.}
\label{fig:trc}
\end{figure*}

The TRC is a training-free, frame-level scalar in $[0,1]$ where 1 means fully reliable. Let $\hat I_T$ be the conditioned thermal frame and $I_R$ the visible frame. Five cues are computed (Fig.~\ref{fig:trc}, Algorithm~1).

\textit{Contrast.} With $H$ the Shannon entropy in bits of the 256-bin histogram of $\hat I_T$ \cite{shannon},
\begin{equation}
S_C = H/8.
\end{equation}
A flat or saturated frame concentrates its histogram in few bins, so $H$ and $S_C$ are low.

\textit{Gradient (sharpness).} With $v=\Var\!\left[\nabla^2 \hat I_T\right]$ the variance of the Laplacian, a standard focus measure \cite{pech},
\begin{equation}
S_G = \frac{v}{v+K},\qquad K=15.
\end{equation}
Blur and defocus remove high-frequency energy, so $v$ and $S_G$ fall. $K$ is a fixed constant that was set once before the experiments reported here.

\textit{Saturation.} With $N_c$ the number of pixels within two grey levels of 0 or 255 and $N$ the total,
\begin{equation}
S_S = 1 - N_c/N.
\end{equation}

\textit{Mutual information.} With $\mathcal{I}$ the mutual information of the joint $32\times 32$ histogram of $\hat I_T$ and $I_R$ \cite{viola},
\begin{equation}
S_{MI} = \frac{\mathcal{I}(\hat I_T; I_R)}{\min\{H(\hat I_T), H(I_R)\}}.
\end{equation}
Thermal crossover and misregistration weaken the statistical dependence between the two images.

\textit{Alignment.} With $\lambda = 5$\,px,
\begin{equation}
S_A = e^{-d/\lambda},
\end{equation}
which is 1 for pre-registered pairs ($d=0$) and is also used when $d$ is unavailable.

The five cues are combined as
\begin{equation}
\rT = \clip_{[0,1]}\!\left(\frac{\sum_i w_i S_i}{\sum_i w_i}\right),\qquad w_i = 0.2,
\end{equation}
where a cue that cannot be computed is dropped and the remaining weights are renormalised. The constants $K$, $\lambda$ and $w_i$ are hand-set and shared by all experiments below.

\begin{algorithm}[t]
\caption{Thermal Reliability Coefficient for one frame pair}
\label{alg:trc}
\begin{algorithmic}[1]
\Require thermal frame $I_T$, visible frame $I_R$, alignment error $d$ (optional)
\Ensure reliability $\rT\in[0,1]$
\State $(\hat I_T,\hat I_R)\gets$ align, denoise, normalise$(I_T,I_R)$
\State $S_C\gets H(\hat I_T)/8$
\State $S_G\gets v/(v+K)$ with $v=\Var[\nabla^2\hat I_T]$
\State $S_S\gets 1-N_c/N$
\State $S_{MI}\gets \mathcal{I}(\hat I_T;\hat I_R)/\min\{H(\hat I_T),H(\hat I_R)\}$
\State $S_A\gets e^{-d/\lambda}$ if $d$ is available else $1$
\State $\rT\gets \clip_{[0,1]}\!\left(\sum_i w_iS_i/\sum_i w_i\right)$
\State \Return $\rT$
\end{algorithmic}
\end{algorithm}

\subsection{TRC-Gated Dual-Branch Fusion Detector}
\begin{figure*}[!t]
\centering
\includegraphics[width=\textwidth]{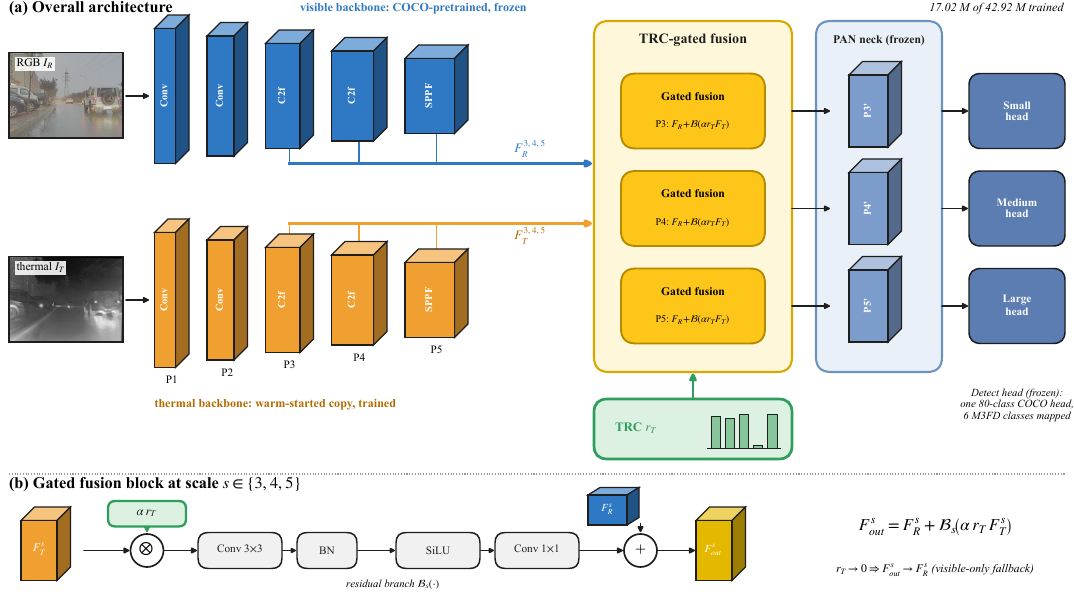}
\caption{TRC-gated dual-branch fusion detector. (a) The visible branch (blue), the PAN neck and the head are frozen; the thermal branch (orange) and the gated fusion blocks (yellow) are trained. Slab height indicates spatial size and thickness the channel count. (b) The gated fusion block applied at each of the three scales.}
\label{fig:fusion}
\end{figure*}

The detector (Fig.~\ref{fig:fusion}) is built on YOLOv8m \cite{yolov8}. The visible branch is the COCO-pretrained backbone (layers 0--9) and is frozen. The thermal branch is a copy of the same backbone whose first convolution is replaced by a one-channel stem initialised with the channel mean of the pretrained RGB kernels, and it is trained. Features at the three scales P3, P4 and P5 (strides 8, 16, 32; 192, 384 and 576 channels) are fused as
\begin{equation}
F^{s}_{out} = F^{s}_{R} + \mathcal{B}_s\!\left(\alpha\,\rT\,F^{s}_{T}\right),\qquad s\in\{3,4,5\},
\end{equation}
where $\mathcal{B}_s$ is a $3\times 3$ convolution, batch normalisation, SiLU and a $1\times 1$ convolution, and $\alpha$ is a learnable scalar initialised to 1. The last convolution is initialised with small weights (standard deviation 0.01) and zero bias, so at initialisation $F^{s}_{out}\approx F^{s}_{R}$ and the pretrained neck receives its usual input. When $\rT\to 0$ the fused feature returns to the visible one.

The pretrained PAN neck (layers 10--21) and the Detect head (layer 22) are frozen. In total 17.02\,M of 42.92\,M parameters are trained: the thermal backbone (11.85\,M) and the fusion blocks (5.16\,M).

The head keeps its 80-class COCO layout. The six M3FD classes are mapped to existing COCO channels (person$\to$person, car$\to$car, bus$\to$bus, truck$\to$truck, motorcycle$\to$motorcycle) and lamp is mapped to the COCO traffic light channel, the nearest pretrained one. The standard YOLOv8 detection loss (box 7.5, classification 0.5, DFL 1.5) is restricted to the mapped channels, and only these channels are decoded at inference.

\subsection{TRC-Gated Multi-Object Tracking}
\begin{figure*}[!t]
\centering
\includegraphics[width=\textwidth]{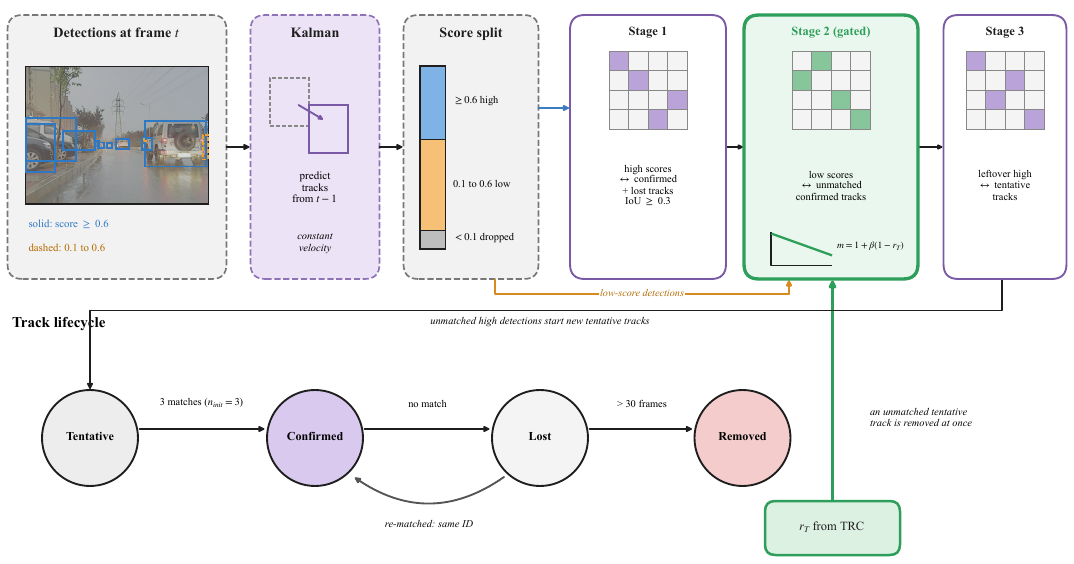}
\caption{TRC-gated tracking. Stage 2 (green) is the only stage that uses the frame reliability $\rT$: its cost is multiplied by $m=1+\beta(1-\rT)$. Grids illustrate the cost matrices. The lower part shows the track lifecycle.}
\label{fig:tracking}
\end{figure*}

The tracker (Fig.~\ref{fig:tracking}) follows ByteTrack \cite{bytetrack}. Each track carries a constant-velocity Kalman state $(c_x,c_y,a,h)$ and its velocities \cite{kalman}. Per frame, all tracks are predicted and detections are split by score into high ($\ge 0.6$), low (0.1--0.6) and discarded ($<0.1$). Assignment uses the cost $c_{ij}=1-\mathrm{IoU}$ and the Hungarian algorithm \cite{kuhn} in three stages: (1) high detections against confirmed and lost tracks, IoU $\ge 0.3$; (2) low detections against the remaining confirmed tracks; (3) leftover high detections against tentative tracks. Unmatched high detections start tentative tracks; a track is confirmed after $n_{init}=3$ matches, becomes lost when unmatched and is removed after 30 frames.

The TRC enters only in stage 2:
\begin{equation}
c'_{ij} = c_{ij}\left[1+\beta\left(1-\rT^{(j)}\right)\right],\qquad \beta=0.5,
\end{equation}
so at $\rT=1$ the tracker is ByteTrack and at low $\rT$ a low-confidence box must agree more closely with the motion prediction. The acceptance threshold in stage 2 is IoU $\ge 0.5$ at $\rT=1$. A missing or non-finite TRC is treated as 1.

\subsection{Behaviour Analysis}
Behaviour features are computed from the matched observations of each confirmed track after smoothing the path with a 5-sample moving average. Speed is measured over a 5-observation window in box heights per frame, so thresholds do not depend on camera resolution. Path efficiency is the net displacement divided by the path length. Direction reversals count consecutive moving samples whose headings differ by more than $90^\circ$. The analyser raises the following flags for the person and motorcycle classes: loitering (a trailing streak of at least 150 frames below 0.01 box heights per frame), erratic path (path efficiency below 0.4 after at least 50 observations), pacing (at least two reversals within the last 20 velocity samples) and restricted-zone dwell (at least 25 frames inside a configured polygon). A low-TRC track flag (mean TRC below 0.5) is raised for every class so that weak evidence is shown rather than hidden. All thresholds are explicit configuration values and nothing is learned.

\section{Experimental Setup}
\begin{figure*}[!t]
\centering
\includegraphics[width=\textwidth]{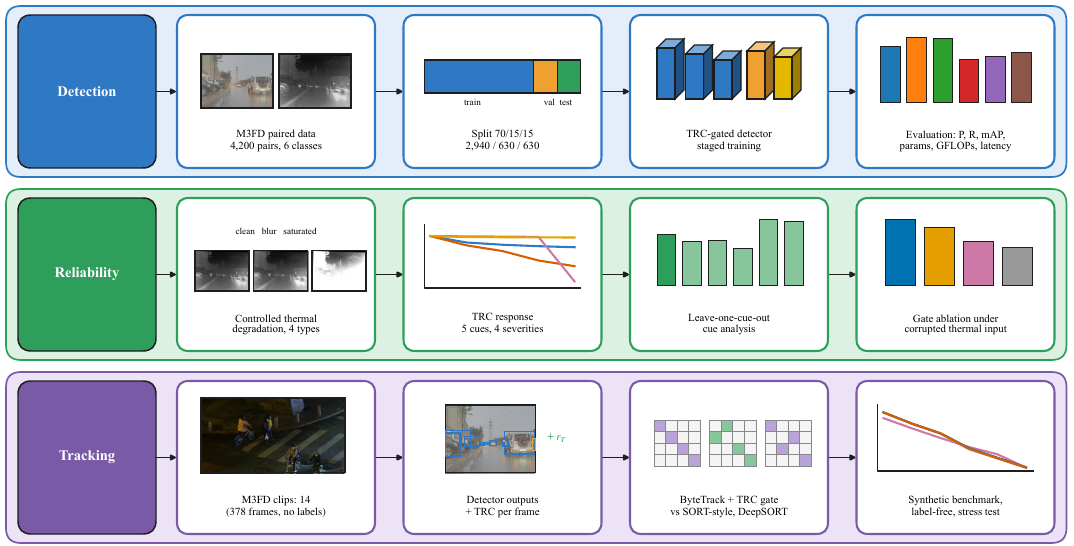}
\caption{Experimental workflow. Detection is evaluated on the M3FD test split, reliability with controlled thermal degradations and a gate ablation, and tracking on synthetic sequences with exact ground truth and on the assembled clips without track labels.}
\label{fig:workflow}
\end{figure*}

Fig.~\ref{fig:workflow} summarises the three evaluation lanes: detection, reliability and tracking.

\subsection{Datasets}
\begin{figure*}[!t]
\centering
\includegraphics[width=\textwidth]{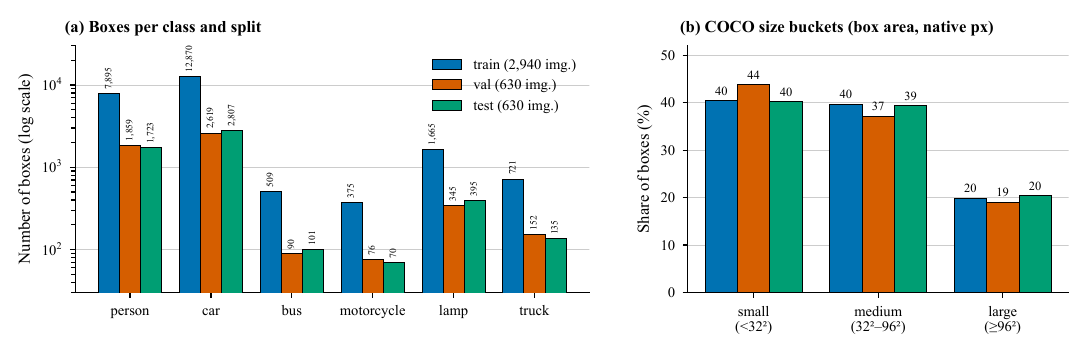}
\caption{M3FD statistics for the split used here. (a) Boxes per class (log scale). (b) Share of boxes in the three COCO size buckets, computed from the box area in native pixels.}
\label{fig:dataset}
\end{figure*}

M3FD \cite{m3fd} provides 4,200 pixel-aligned visible-infrared pairs of size $1024\times768$ with bounding boxes for six classes: person, car, bus, motorcycle, truck and lamp. The dataset has no official split, so all pairs were pooled and divided at random with seed 42 into 2,940 training, 630 validation and 630 test pairs (70/15/15). Table~\ref{tab:data} lists the statistics and Fig.~\ref{fig:dataset} shows the class and size distributions. The classes are strongly imbalanced: the test split has only 70 motorcycle and 101 bus boxes, so the AP of these classes is uncertain. About 40\% of all boxes are small (area below $32^2$\,px$^2$).

\begin{table}[!t]
\caption{Statistics of the Data Used. Boxes Are Listed as Train/Val/Test.}
\label{tab:data}
\centering
\footnotesize
\setlength{\tabcolsep}{3pt}
\begin{tabular}{lrrrrr}
\toprule
Data & Pairs & Classes & Size & Input & Split (pairs)\\
\midrule
M3FD & 4,200 & 6 & $1024\times768$ & $640\times640$ & 2,940/630/630\\
M3FD clips & 14 & none & $1024\times768$ & --- & ---\\
\bottomrule
\end{tabular}

\vspace{2pt}
\raggedright\footnotesize M3FD boxes: 24,035 / 5,141 / 5,231 (total 34,407)
\end{table}

\textit{Tracking clips.} Neighbouring M3FD files often come from the same scene: the boxes of frame $i$ and frame $i+1$ overlap heavily. Runs of consecutive files were grouped into clips whenever at least half of the boxes of one frame matched a box of the next at IoU above 0.5. This gives 2,409 runs, of which 52 have 10 or more frames; the longest has 42. The tracking tests use the 14 clips with at least 18 frames (378 frames in total, 5 of them clips at night), a threshold that was fixed before any tracker was run. The clips are stored as MP4 files at the $1024\times768$ resolution of M3FD. They carry no track annotations, and because the frames are still images their timing is that of the assembly, not of a camera. The clips are drawn from all 4,200 pairs, so 70\% of their frames (264 of 378) belong to the training split of the detector, 13\% to validation and 17\% to test. This does not affect the comparison between trackers, which receive identical detections, but the detections on these clips are better than on unseen footage, and no detection accuracy is claimed for them.

\subsection{Experimental Environment and Settings}
All experiments ran on a single laptop (Table~\ref{tab:env}). Training settings are listed in Table~\ref{tab:train}. Because of the 6\,GB GPU and of interruptions, the training run departs from a standard schedule in several ways, which are disclosed here rather than hidden:
\begin{itemize}
\item Mixed precision (AMP) was enabled and mosaic augmentation was not implemented; horizontal flip, rotation, colour jitter, random resized crop and thermal Gaussian noise were used instead.
\item Inputs are resized to $640\times640$ without letterboxing, which stretches the 4:3 frames slightly.
\item For the first 35 epochs a mapping error meant that only the person class was supervised. It was corrected at epoch 35, and the lamp class was added at epoch 42. The validation mAP therefore changes definition at these epochs (Fig.~\ref{fig:training}).
\item Training was resumed from checkpoints many times. The logged learning rate shows small downward steps at epochs 35, 37, 42 and 46, so the schedule is only approximately cosine. From epoch 101 the effective batch of 16 was obtained with two micro-batches of 8.
\item Only stage 1 (thermal branch and fusion blocks trained) was run, with one run and one seed, so no run-to-run variance is available.
\end{itemize}

\begin{table}[!t]
\caption{Experimental Environment.}
\label{tab:env}
\centering
\footnotesize
\begin{tabular}{ll}
\toprule
Item & Value\\
\midrule
CPU & AMD Ryzen 7 5800H (8C/16T)\\
RAM & 15.3\,GB\\
GPU & RTX 3060 Laptop, 6\,GB\\
Operating system & Windows 11\\
Python / PyTorch \cite{pytorch} & 3.11.16 / 2.3.1 (CUDA 12.1)\\
Ultralytics \cite{yolov8} & 8.4.153\\
Evaluation & pycocotools 2.0.11\\
\bottomrule
\end{tabular}
\end{table}

\begin{table}[!t]
\caption{Training Settings of the Detector.}
\label{tab:train}
\centering
\footnotesize
\begin{tabular}{p{0.34\columnwidth}p{0.58\columnwidth}}
\toprule
Parameter & Value\\
\midrule
Epochs & 150 planned; early stop at 131 (patience 15)\\
Selected checkpoint & epoch 116 (best validation mAP@50)\\
Batch size & $8\times2$ accumulation = 16 (effective)\\
Input size & $640\times640$\\
Optimiser & SGD (momentum 0.937, weight decay $5\times10^{-4}$)\\
Learning rate & 0.01, cosine \cite{sgdr} to $10^{-4}$, no warm-up\\
Gradient clipping & norm 10\\
Loss gains & box 7.5, cls 0.5, DFL 1.5\\
Precision & mixed (AMP)\\
Augmentation & flip 0.5, rotation $\pm15^\circ$, resized crop (0.6--1.0), colour jitter, thermal noise\\
Parameters (trained / total) & 17.02\,M / 42.92\,M\\
Seed & 42 (one run)\\
\bottomrule
\end{tabular}
\end{table}

\subsection{Evaluation Metrics}
\textit{Detection.} Precision, recall, mAP@50 and mAP@50:95 follow the COCO protocol \cite{coco} using pycocotools at the deployment thresholds (confidence 0.25, NMS IoU 0.45); $F_1=2PR/(P+R)$. Efficiency is reported as parameters, GFLOPs (counted by the thop profiler as $2\times$MACs, an approximation), latency at batch size 1 on the GPU, and FPS $=1000/$latency.

\textit{Reliability.} For each degradation and severity, the mean and standard deviation of $\rT$ over frames; the Spearman rank correlation between severity and $\rT$ per frame, averaged over frames; the share of frames whose $\rT$ decreases monotonically with severity; and the share of frames with $\rT<0.5$. For the leave-one-cue-out analysis the TRC is recomputed from the stored cue scores with one cue removed and the weights renormalised.

\textit{Tracking.} On synthetic sequences with exact ground truth: HOTA, DetA and AssA \cite{hota}, MOTA \cite{clearmot}, IDF1 \cite{idf1}, identity switches (IDSW), false positives and false negatives. Paired differences between trackers over the same sequences use a bootstrap 95\% confidence interval \cite{efron} and the Wilcoxon signed-rank test \cite{wilcoxon}. The clips have no labels, so only label-free proxies are reported: the number of tracks, persistence (mean track length divided by the clip length), tracks per concurrent target (number of tracks divided by the mean number of tracked objects per frame), coverage of confident detections, the rate of implausible centre jumps, suspected identity fragmentation and tracker time per frame. The unit of analysis is the clip: every statistic is first computed per clip, and paired differences between trackers over the 14 clips use a bootstrap 95\% confidence interval (10,000 resamples of clips) and the Wilcoxon signed-rank test. These proxies cannot show that a track is correct.

\subsection{Reliability and Tracking Protocols}
\textit{Controlled thermal degradation.} A deterministic subsample of 150 test pairs (every fourth pair) is degraded on the raw thermal frame, before conditioning, at severities $\delta\in\{0.25,0.5,0.75,1\}$: blur (Gaussian kernel 7, 11, 15, 21), saturation (gain $1+3\delta$ followed by a $9\times9$ bloom blur), flat (blend of the frame toward its mean by $\delta$, giving a dead frame at $\delta=1$) and misalignment (horizontal shift of $24\delta$ pixels).

\textit{Gate ablation.} The trained checkpoint is evaluated on all 630 test pairs, clean and with the thermal stream corrupted (blur or saturation at full severity, chosen per image from the file name). Only the reliability value passed to the fusion blocks changes: adaptive (measured $\rT$), fixed ($\rT=0.782$, the mean clean value), full thermal ($\rT=1$) and thermal off ($\rT=0$, so that $F_{out}=F_R$). This is an inference-time substitution on one set of weights, not a comparison of separately trained models.

\textit{Tracking validation} has four layers: (1) unit tests of the tracker, gate, behaviour and metric code (28 and 17 checks); (2) 30 seeded synthetic sequences with exact ground truth, whose simulated detector degrades when $\rT$ dips (the coupled condition), with two controls in which the gate has no reason to help (uninformative: constant $\rT$ given to the tracker; no degradation: $\rT$ dips but detections are unaffected); (3) label-free metrics on the 14 M3FD clips; (4) a stress test in which, in every one of these clips, the thermal stream is blurred or saturated (two separate runs) in the middle third of the clip while RGB is untouched, measuring how detector output depends on $\rT$ and how stable each tracker's identities remain (each degraded run scored against the same tracker's clean-run tracks inside the degraded segment, so this is self-consistency, not accuracy). Every clip is used, with no selection, and the two corruption types of a clip are averaged before analysis so that the 14 clips remain the independent units. All trackers receive identical detections and TRC values. The compared trackers are a SORT-style tracker (ByteTrack without the low-confidence stage), DeepSORT with its appearance embedder disabled (motion only), ByteTrack, and ByteTrack with the TRC gate.

\section{Results and Discussion}

\subsection{Detection on M3FD}
\begin{figure*}[!t]
\centering
\includegraphics[width=\textwidth]{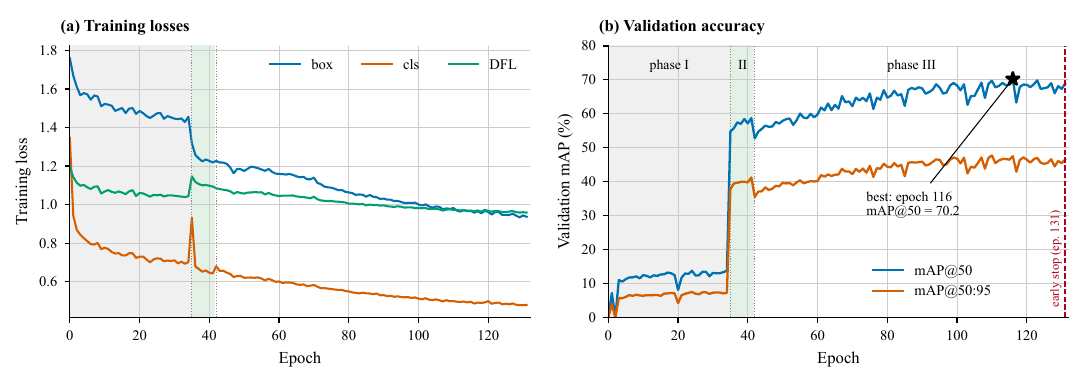}
\caption{Training history of the selected run. Shaded regions mark phase I (epochs 0--34, only the person class supervised), phase II (35--41, five classes) and phase III (42--131, six classes); the validation mAP changes definition at the phase boundaries. The star marks the selected epoch and the dashed line the early stop.}
\label{fig:training}
\end{figure*}

Fig.~\ref{fig:training} shows the training history. The selected checkpoint (epoch 116) reaches 70.2\% mAP@50 and 47.5\% mAP@50:95 on the validation split, and Table~\ref{tab:detection} reports its test-split results: 87.6\% precision, 79.4\% recall ($F_1=83.3\%$), 70.0\% mAP@50 and 47.8\% mAP@50:95. Validation and test agree closely, so the checkpoint selection did not overfit the validation split. Car (84.3\% mAP@50) and bus (82.5\%) are detected best. Motorcycle (55.7\%), lamp (60.1\%) and truck (64.3\%) are weakest, and for motorcycle and lamp recall (58.6\% and 62.8\%) is much lower than precision (78.8\% and 84.6\%), that is, the detector is conservative on rare and small objects. Lamp has to reuse the traffic-light channel, and motorcycle has only 375 training boxes.

\begin{table}[!t]
\caption{Detection Results of TriNetra on the M3FD Test Split (630 Pairs, One Run). The 95\% Confidence Intervals Resample the Test Images (1,000 Bootstrap Resamples). Efficiency Was Measured With Batch Size 1 on the RTX 3060 Laptop GPU.}
\label{tab:detection}
\centering
\footnotesize
\setlength{\tabcolsep}{3pt}
\begin{tabular}{lrrcr}
\toprule
Class & P (\%) & R (\%) & mAP@50 [95\% CI] & mAP@50:95\\
\midrule
All & 87.6 & 79.4 & 70.0 [66.7, 73.4] & 47.8\\
Person & 85.2 & 75.3 & 72.9 [69.9, 75.8] & 44.3\\
Car & 90.0 & 85.0 & 84.3 [82.3, 85.5] & 62.0\\
Bus & 85.9 & 84.2 & 82.5 [74.5, 89.1] & 70.4\\
Motorcycle & 78.8 & 58.6 & 55.7 [44.4, 67.4] & 31.4\\
Lamp & 84.6 & 62.8 & 60.1 [54.5, 66.3] & 30.6\\
Truck & 79.0 & 69.6 & 64.3 [54.8, 74.1] & 48.1\\
\midrule
\multicolumn{5}{l}{Params 42.92\,M; GFLOPs 130.0 (approx.)}\\
\multicolumn{5}{l}{Latency 139.8\,ms (batch 1); 7.2 FPS}\\
\bottomrule
\end{tabular}
\end{table}

\begin{figure}[!t]
\centering
\includegraphics[width=0.95\columnwidth]{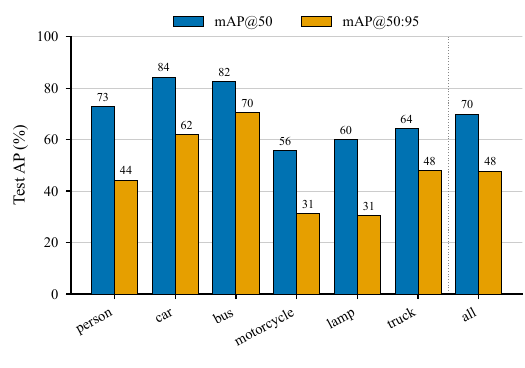}
\caption{Per-class test AP of TriNetra on M3FD.}
\label{fig:perclass}
\end{figure}

\textit{Comparison with published M3FD results.} Table~\ref{tab:compare} places the result next to the M3FD numbers that GEM-YOLO reports for several dual-stream detectors \cite{gemyolo}. The comparison is indicative only. That study splits the same 4,200 pairs 7:2:1 (420 test pairs), trains its detectors from scratch, averages five runs, and uses small models with 3.4--7.5\,M parameters (RT-DETR: 66.3\,M), whereas TriNetra uses one run, a different 70/15/15 split, and a frozen COCO-pretrained visible path with 42.9\,M parameters. Under these caveats TriNetra has the highest mAP@50:95 in the table (47.8\% against 42.2--44.7\%) and an mAP@50 (70.0\%) close to the best reported value (70.3\%), but it is far larger and slower (7.2 FPS on a laptop GPU, so it is not a real-time detector in this setting). Its accuracy therefore should not be read as an efficiency claim.

\begin{table}[!t]
\caption{Indicative Comparison on M3FD (P, R, mAP in \%; Parameters in Millions). Values for Other Methods Are Copied From \cite{gemyolo} (Mean of Five Runs, 7:2:1 Split, Trained From Scratch); TriNetra Is One Run on a 70/15/15 Split. DS: Dual-Stream. Precision and Recall Are Not Directly Comparable Across Studies.}
\label{tab:compare}
\centering
\scriptsize
\setlength{\tabcolsep}{2.5pt}
\begin{tabular}{lrrrrrr}
\toprule
Method & P & R & mAP@50 & mAP@50:95 & Params & GFLOPs\\
\midrule
YOLOv8n (DS) & 82.0 & 62.8 & 70.3 & 44.7 & 4.36 & 11.34\\
YOLOv11n (DS) & 82.6 & 57.5 & 66.5 & 42.6 & 3.79 & 9.31\\
MBNet & 80.0 & 61.2 & 68.2 & 43.0 & 6.8 & 16.2\\
ProbEn & 80.8 & 61.7 & 68.8 & 43.3 & 7.5 & 17.4\\
RT-DETR (DS) & 79.2 & 60.5 & 67.2 & 42.2 & 66.34 & 194.03\\
GEM-YOLO \cite{gemyolo} & 79.2 & 61.8 & 69.0 & 43.8 & 3.44 & 7.58\\
TriNetra (ours) & 87.6 & 79.4 & 70.0 & 47.8 & 42.92 & 130.0\\
\bottomrule
\end{tabular}
\end{table}

\subsection{Behaviour of the Thermal Reliability Coefficient}
\begin{figure*}[!t]
\centering
\includegraphics[width=\textwidth]{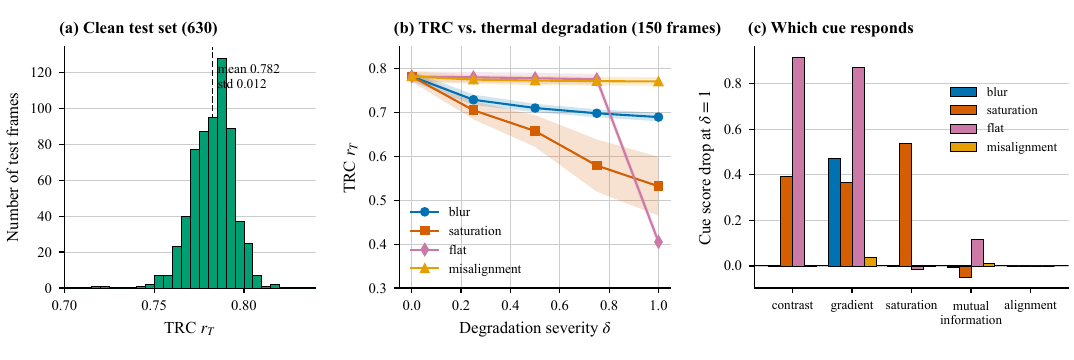}
\caption{TRC analysis on M3FD. (a) Distribution over the 630 clean test pairs. (b) TRC against degradation severity (mean $\pm$ one standard deviation over 150 pairs). (c) Drop of each cue score at full severity, showing which cue reacts to which degradation.}
\label{fig:trcplots}
\end{figure*}

\textit{Clean data.} Over all 630 clean test pairs the TRC has mean 0.782, standard deviation 0.012, minimum 0.720 and maximum 0.818 (Fig.~\ref{fig:trcplots}a); no frame falls below 0.5. Table~\ref{tab:cues} gives the individual cues. Contrast (0.94), sharpness (0.87) and saturation (0.99) are high and stable, alignment is 1.00 because the pairs are pre-registered, and the mutual-information cue is low (0.115), because thermal and visible intensities are only weakly dependent. On clean M3FD data the TRC is therefore almost constant, and the gate acts nearly as a fixed scale.

\begin{table}[!t]
\caption{Cue Scores and TRC on the 630 Clean M3FD Test Pairs (Mean $\pm$ Standard Deviation).}
\label{tab:cues}
\centering
\footnotesize
\begin{tabular}{lrr}
\toprule
Score & Mean & Std\\
\midrule
Contrast $S_C$ & 0.938 & 0.023\\
Gradient $S_G$ & 0.873 & 0.046\\
Saturation $S_S$ & 0.987 & 0.003\\
Mutual information $S_{MI}$ & 0.115 & 0.059\\
Alignment $S_A$ & 1.000 & 0.000\\
TRC $\rT$ & 0.782 & 0.012\\
\bottomrule
\end{tabular}
\end{table}

\textit{Response to degradation.} Table~\ref{tab:sweep} and Fig.~\ref{fig:trcplots}b show the TRC under the four degradations. Blur lowers the TRC monotonically in every frame, but only to 0.690 at full severity, because sharpness is the one cue that responds (Fig.~\ref{fig:trcplots}c). Saturation lowers it more (0.532 at full severity, with 30\% of frames below 0.5), through the contrast, sharpness and saturation cues together. A flat frame is detected only when it is fully flat: partial flattening barely changes the TRC (0.776 at $\delta=0.75$) because the normalisation step restores the contrast, and at $\delta=1$ the TRC drops to 0.405 in all frames. Misalignment up to 24\,px changes the TRC by only 0.011: the alignment cue is constant when no alignment error is estimated, and the mutual-information cue is too low and too flat to react. The TRC therefore does not detect misregistration in the present configuration.

\begin{table}[!t]
\caption{Mean TRC Over 150 Test Pairs (Clean Value 0.782) at Severity $\delta$; the Standard Deviation Is 0.009--0.012 Except for Saturation (0.021--0.067, See Fig.~\ref{fig:trcplots}b). $\rho$: Mean Per-Frame Spearman Correlation Between $\delta$ and TRC; Mono.: Share of Frames Whose TRC Never Increases With $\delta$; $<$0.5: Share of Frames Below 0.5 at $\delta=1$.}
\label{tab:sweep}
\centering
\scriptsize
\setlength{\tabcolsep}{2.5pt}
\begin{tabular}{lrrrrrrr}
\toprule
 & \multicolumn{4}{c}{$\delta$} & & & \\
\cmidrule(lr){2-5}
Degradation & 0.25 & 0.5 & 0.75 & 1.0 & $\rho$ & Mono. & $<$0.5\\
\midrule
Blur & .729 & .710 & .698 & .690 & $-1.00$ & 100\% & 0\%\\
Saturation & .705 & .658 & .579 & .532 & $-1.00$ & 96\% & 30\%\\
Flat & .780 & .778 & .776 & .405 & $-0.74$ & 55\% & 100\%\\
Misalignment & .775 & .773 & .772 & .771 & $-0.93$ & 84\% & 0\%\\
\bottomrule
\end{tabular}
\end{table}

\textit{Cue ablation.} Table~\ref{tab:loco} removes one cue at a time. Without the gradient cue the response to blur vanishes ($-0.002$), so sharpness is the only blur detector. The contrast and saturation cues carry the response to saturation and flat frames. The alignment and mutual-information cues add an almost constant offset on this pre-registered dataset, and removing them increases the separation between clean and degraded frames (for saturation from 0.250 to 0.325 and 0.313; for flat frames from 0.377 to 0.443 and 0.472). With equal weights these two cues compress the dynamic range of the TRC, which points to re-weighting them, or to estimating the alignment error, as a way to widen the range.

\begin{table}[!t]
\caption{Leave-One-Cue-Out Analysis on 150 Test Pairs. ``Level'' Is the Mean Clean TRC; the Other Columns Are the Drop of the Mean TRC From Clean to $\delta=1$ (Larger Means Better Separation).}
\label{tab:loco}
\centering
\footnotesize
\begin{tabular}{lrrrr}
\toprule
Variant & Level & Blur & Saturation & Flat\\
\midrule
All five cues & 0.782 & 0.093 & 0.250 & 0.377\\
Without contrast & 0.744 & 0.116 & 0.214 & 0.243\\
Without gradient & 0.760 & $-0.002$ & 0.221 & 0.254\\
Without saturation & 0.731 & 0.116 & 0.178 & 0.475\\
Without mutual information & 0.949 & 0.117 & 0.325 & 0.443\\
Without alignment & 0.728 & 0.116 & 0.313 & 0.472\\
\bottomrule
\end{tabular}
\end{table}

\subsection{Gate Ablation Under Corrupted Thermal Input}
\begin{figure*}[!t]
\centering
\includegraphics[width=\textwidth]{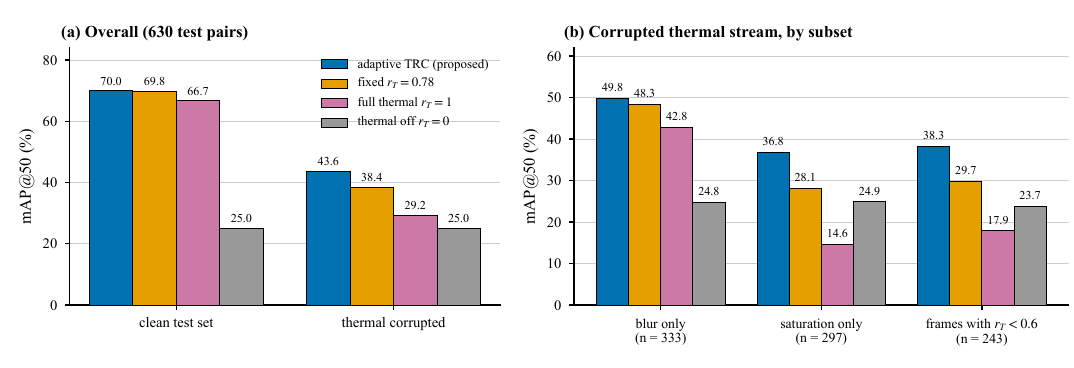}
\caption{Gate ablation on the M3FD test split with identical weights. (a) Overall mAP@50 on clean and corrupted thermal input. (b) mAP@50 on the corrupted stream by subset.}
\label{fig:gate}
\end{figure*}

Table~\ref{tab:gate} and Fig.~\ref{fig:gate} compare the four gate modes on the same weights. On clean data the adaptive and fixed gates are equal (70.0\% and 69.8\% mAP@50), which is expected because the measured TRC is almost constant there. Admitting the thermal features at full strength lowers accuracy by 3.3 points (66.7\%), showing that the network was trained around the operating value $\rT\approx0.78$. With the thermal stream corrupted (mean TRC 0.616), the adaptive gate keeps 43.6\% mAP@50 against 38.4\% for the fixed gate, 29.2\% for full thermal and 25.0\% for the closed gate. The advantage of adaptation is largest on saturated frames (36.8\% against 28.1\%, +8.7 points) and on frames with $\rT<0.6$ (38.3\% against 29.7\%), and small for blur (49.8\% against 48.3\%), where the TRC reacts weakly (Table~\ref{tab:sweep}). This is the strongest evidence in this study that the score carries useful information.

\textit{Uncertainty.} To separate real differences from sampling noise, the 630 test pairs were resampled with replacement 1,000 times, with the same resample applied to every gate mode so that the differences are paired (Tables~\ref{tab:gateci} and~\ref{tab:gatediff}). The 95\% intervals of the overall mAP@50 extend about 3.5 points to either side of the estimate, so a gap of a few tenths between two separately evaluated numbers cannot be interpreted. Even so, the main result is clear. On clean data the adaptive and fixed gates do not differ ($+0.1$ points, 95\% CI $[-0.2,+0.2]$). On corrupted thermal input the adaptive gate is better than the fixed gate by $+5.1$ points ($[+3.6,+6.5]$ for mAP@50 and $[+2.4,+4.3]$ for mAP@50:95), and the interval excludes zero in all three subsets: saturation ($+8.8$, $[+6.1,+12.2]$), frames with $\rT<0.6$ ($+8.6$, $[+5.4,+12.1]$) and blur ($+1.5$, $[+0.9,+2.3]$). Against full thermal admission the adaptive gate is better by $+3.3$ points on clean data ($[+2.1,+4.6]$) and by $+14.4$ points on corrupted data ($[+12.1,+16.6]$). The intervals describe uncertainty over test images only; they do not cover run-to-run variation of training, which was not measured. The bootstrap evaluator reproduces the COCO evaluation of Table~\ref{tab:gate} to the first decimal for every entry except one (fixed gate on saturated frames, 28.0 against 28.1), so the re-run on a CPU reproduced the original GPU results.

Two limits apply. First, the adaptive gate protects only partly: accuracy still falls from 70.0\% to 43.6\%, because a corrupted frame with $\rT\approx0.5$ still passes half of its damaged features. Second, the closed gate (25.0\%) is the frozen COCO-pretrained visible path with no M3FD fine-tuning, so it should not be read as a trained RGB-only detector.

\begin{table*}[!t]
\caption{Gate Ablation With Identical Weights on the M3FD Test Split (630 Pairs): mAP@50 / mAP@50:95 in \%. Corrupted: Blur or Saturation at Full Severity on the Thermal Stream; the Subsets Are Blur (333 Pairs), Saturation (297) and $\rT<0.6$ (243).}
\label{tab:gate}
\centering
\footnotesize
\begin{tabular}{lccccc}
\toprule
Gate mode & Clean & Corrupted & Blur & Saturation & $\rT<0.6$\\
\midrule
Adaptive TRC (proposed) & 70.0 / 47.8 & 43.6 / 30.6 & 49.8 / 35.5 & 36.8 / 25.6 & 38.3 / 27.2\\
Fixed $\rT=0.782$ & 69.8 / 47.8 & 38.4 / 27.3 & 48.3 / 34.4 & 28.1 / 20.0 & 29.7 / 21.6\\
Full thermal $\rT=1$ & 66.7 / 44.7 & 29.2 / 20.7 & 42.8 / 30.1 & 14.6 / 10.8 & 17.9 / 13.6\\
Thermal off $\rT=0$ & 25.0 / 14.4 & 25.0 / 14.4 & 24.8 / 14.6 & 24.9 / 14.5 & 23.7 / 14.0\\
\bottomrule
\end{tabular}
\end{table*}

\begin{table}[!t]
\caption{Overall mAP@50 (\%) of the Four Gate Modes With 95\% Bootstrap Confidence Intervals (630 Test Pairs, 1,000 Image Resamples), and the Paired Difference Between the Adaptive Gate and Each Other Mode on Corrupted Thermal Input.}
\label{tab:gateci}
\centering
\scriptsize
\setlength{\tabcolsep}{2.5pt}
\begin{tabular}{lccc}
\toprule
Gate mode & Clean & Corrupted & Adaptive $-$ mode (corrupted)\\
\midrule
Adaptive TRC & 70.0 [66.7, 73.4] & 43.6 [40.0, 47.2] & ---\\
Fixed $\rT=0.782$ & 69.8 [66.7, 73.3] & 38.4 [34.9, 42.2] & +5.1 [+3.6, +6.5]\\
Full thermal $\rT=1$ & 66.7 [63.2, 70.0] & 29.2 [25.8, 32.7] & +14.4 [+12.1, +16.6]\\
Thermal off $\rT=0$ & 25.0 [22.7, 27.2] & 25.0 [22.7, 27.3] & +18.6 [+15.5, +22.2]\\
\bottomrule
\end{tabular}
\end{table}

\begin{table}[!t]
\caption{Adaptive Gate Minus Fixed Gate, mAP@50 in Percentage Points, Paired Bootstrap (95\% CI), by Subset of the Test Split.}
\label{tab:gatediff}
\centering
\footnotesize
\begin{tabular}{lrc}
\toprule
Subset & Pairs & Difference [95\% CI]\\
\midrule
Clean thermal & 630 & +0.1 [$-$0.2, +0.2]\\
Corrupted thermal, all & 630 & +5.1 [+3.6, +6.5]\\
\quad blur only & 333 & +1.5 [+0.9, +2.3]\\
\quad saturation only & 297 & +8.8 [+6.1, +12.2]\\
\quad $\rT<0.6$ & 243 & +8.6 [+5.4, +12.1]\\
\bottomrule
\end{tabular}
\end{table}

\subsection{Tracking}
\begin{figure*}[!t]
\centering
\includegraphics[width=0.345\textwidth]{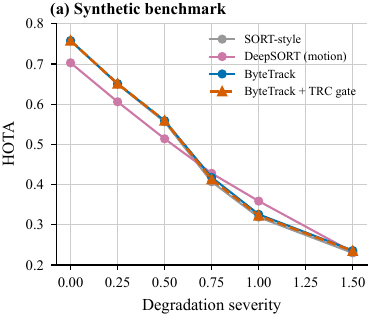}\hfill
\includegraphics[width=0.314\textwidth]{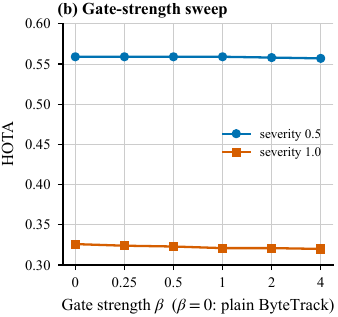}\hfill
\includegraphics[width=0.298\textwidth]{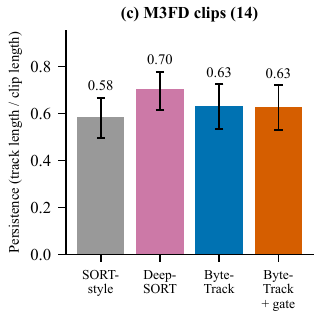}
\caption{Tracking results. (a) HOTA against degradation severity on the synthetic benchmark; the gated tracker and ByteTrack nearly coincide. (b) HOTA against gate strength $\beta$ ($\beta=0$ is plain ByteTrack). (c) Persistence (mean track length divided by clip length) of each tracker on the 14 M3FD clips with identical detections (mean and 95\% CI over clips).}
\label{fig:trkres}
\end{figure*}

\textit{Synthetic benchmark.} Table~\ref{tab:synthetic} and Fig.~\ref{fig:trkres}a give the results on 30 sequences with exact ground truth. At severity 0.5 the TRC-gated tracker and ByteTrack are indistinguishable (HOTA 0.559 for both; paired difference $-0.000$, 95\% CI $[-0.005,+0.005]$), and at severity 1.0 the gated tracker is slightly worse (HOTA 0.323 against 0.326, difference $-0.003$, CI $[-0.005,-0.000]$; MOTA $-0.003$). Both controls show the same pattern ($\Delta$HOTA $\approx-0.001$), and varying the gate strength $\beta$ from 0 to 4 does not help (Fig.~\ref{fig:trkres}b): HOTA at severity 1.0 falls from 0.326 to 0.320. Against the SORT-style tracker the gated tracker is marginally better (HOTA +0.003 and +0.004), and against DeepSORT without appearance it is better at severity 0.5 (+0.044) but worse at severity 1.0 ($-0.037$), so the DeepSORT comparison is not consistent. The small gain over the SORT-style tracker is already present in ByteTrack without the gate.

\textit{M3FD clips.} Table~\ref{tab:clips} compares the trackers on the 14 clips with identical detections. ByteTrack and the gated tracker produce the same 119 tracks, and the paired differences between them are zero for the number of tracks, coverage and jump rate (Wilcoxon $p=1.0$); persistence differs by $-0.004$ (95\% CI $[-0.009,0.000]$). Compared with the SORT-style tracker, ByteTrack keeps identities longer (persistence $+0.047$, CI $[+0.018,+0.082]$, $p=0.008$, higher in 9 of 14 clips and lower in none) and needs fewer tracks per concurrent target ($-0.14$, CI $[-0.23,-0.06]$). Compared with DeepSORT without appearance, ByteTrack covers more of the confident detections ($+0.050$, CI $[+0.014,+0.090]$, $p=0.023$), but DeepSORT has longer tracks on average, so ByteTrack's advantage over it lies in coverage and not in track length. No behaviour flag was raised on any clip, as expected for clips of at most 42 frames when the loitering rule needs 150, so the behaviour analyser is validated only by its unit tests and not by observed events. The tracker alone costs about 0.5\,ms per frame.

\textit{Stress test on the clips.} The assumption behind the gate is that low TRC goes with worse detections. In the middle third of the 14 clips the TRC fell from 0.78 to 0.68 under blur and to 0.50 under saturation (8 of the 14 saturated runs below 0.5). Relative to the clean run, 88\% of the confident detections were still found at TRC between 0.40 and 0.60 and 89\% at TRC between 0.60 and 0.75 (94 and 154 frames), while the detection count per frame fell to 69\% and 75\% and the scores to 76\% and 87\%. Per frame, the rank correlation between the TRC and the share of retained detections is $-0.04$ ($p=0.57$, 252 frames). In this test a lower TRC therefore went with lower scores and fewer detections, but not with more lost objects, and the premise of the gate is only partly supported. The identity stability of the gated tracker was slightly lower than that of ByteTrack (Table~\ref{tab:stress}; IDF1 $-0.015$, 95\% CI $[-0.031,-0.003]$, $p=0.046$; better in 1 clip, worse in 5, equal in 8). The loss comes from the saturated runs ($-0.031$, CI $[-0.061,-0.007]$, $p=0.028$); for blur the difference is zero ($+0.000$, CI $[-0.004,+0.004]$). Both ByteTrack variants are clearly more stable than the SORT-style tracker (IDF1 $+0.059$, CI $[+0.026,+0.095]$) and DeepSORT without appearance ($+0.059$, CI $[+0.024,+0.097]$). The sample is small (14 clips, segments of 6 to 14 frames), so these results show that the gate gives no benefit here; they are not strong evidence that it harms.

\begin{table}[!t]
\caption{Synthetic Tracking Benchmark (30 Sequences, Mean, Coupled Condition) at Degradation Severity 0.5 and 1.0. The Standard Deviation Over Sequences Is 0.04--0.09 for HOTA and 0.03--0.10 for MOTA.}
\label{tab:synthetic}
\centering
\scriptsize
\setlength{\tabcolsep}{2.5pt}
\begin{tabular}{lcccccc}
\toprule
 & \multicolumn{3}{c}{Severity 0.5} & \multicolumn{3}{c}{Severity 1.0}\\
\cmidrule(lr){2-4}\cmidrule(lr){5-7}
Tracker & HOTA & MOTA & IDSW & HOTA & MOTA & IDSW\\
\midrule
SORT-style & .556 & .745 & 12.8 & .319 & .471 & 17.8\\
DeepSORT (motion) & .514 & .702 & 15.9 & .359 & .456 & 27.0\\
ByteTrack & .559 & .754 & 12.4 & .326 & .480 & 17.6\\
ByteTrack + TRC gate & .559 & .751 & 12.5 & .323 & .476 & 17.7\\
\bottomrule
\end{tabular}
\end{table}

\begin{table}[!t]
\caption{Tracker Comparison on the 14 M3FD Clips (378 Frames) With Identical Detections and Label-Free Proxies. Persistence Is Mean Track Length Divided by Clip Length (Mean $\pm$ Std Over Clips); Tracks per Target Is the Number of Tracks Divided by the Mean Number of Tracked Objects per Frame; Jumps Are Implausible Centre Jumps per 1,000 Track-Frames.}
\label{tab:clips}
\centering
\scriptsize
\setlength{\tabcolsep}{2.5pt}
\begin{tabular}{lrcccc}
\toprule
Tracker & Tracks & Persistence & Tracks/target & Coverage & Jumps\\
\midrule
SORT-style & 122 & $0.58\pm0.17$ & 1.92 & 0.792 & 1.3\\
DeepSORT (motion) & 122 & $0.70\pm0.16$ & 1.55 & 0.742 & 0.1\\
ByteTrack & 119 & $0.63\pm0.19$ & 1.79 & 0.793 & 1.3\\
ByteTrack + TRC gate & 119 & $0.63\pm0.19$ & 1.81 & 0.793 & 1.3\\
\bottomrule
\end{tabular}
\end{table}

\begin{table}[!t]
\caption{Identity Stability in the Degraded Middle Third of the 14 M3FD Clips (Blur and Saturation Runs Averaged per Clip). Mean Over Clips With 95\% Bootstrap CI for IDF1. Each Tracker's Degraded Tracks Are Scored Against Its Own Clean-Run Tracks.}
\label{tab:stress}
\centering
\scriptsize
\setlength{\tabcolsep}{2.5pt}
\begin{tabular}{lcccc}
\toprule
Tracker & IDF1 [95\% CI] & HOTA & AssA & IDSW\\
\midrule
SORT-style & .690 [.517, .839] & .639 & .730 & 0.32\\
DeepSORT (motion) & .690 [.498, .860] & .640 & .709 & 0.71\\
ByteTrack & .749 [.562, .904] & .699 & .777 & 0.32\\
ByteTrack + TRC gate & .734 [.551, .887] & .686 & .769 & 0.32\\
\bottomrule
\end{tabular}
\end{table}

\subsection{Qualitative Results}
\begin{figure*}[!t]
\centering
\includegraphics[width=\textwidth]{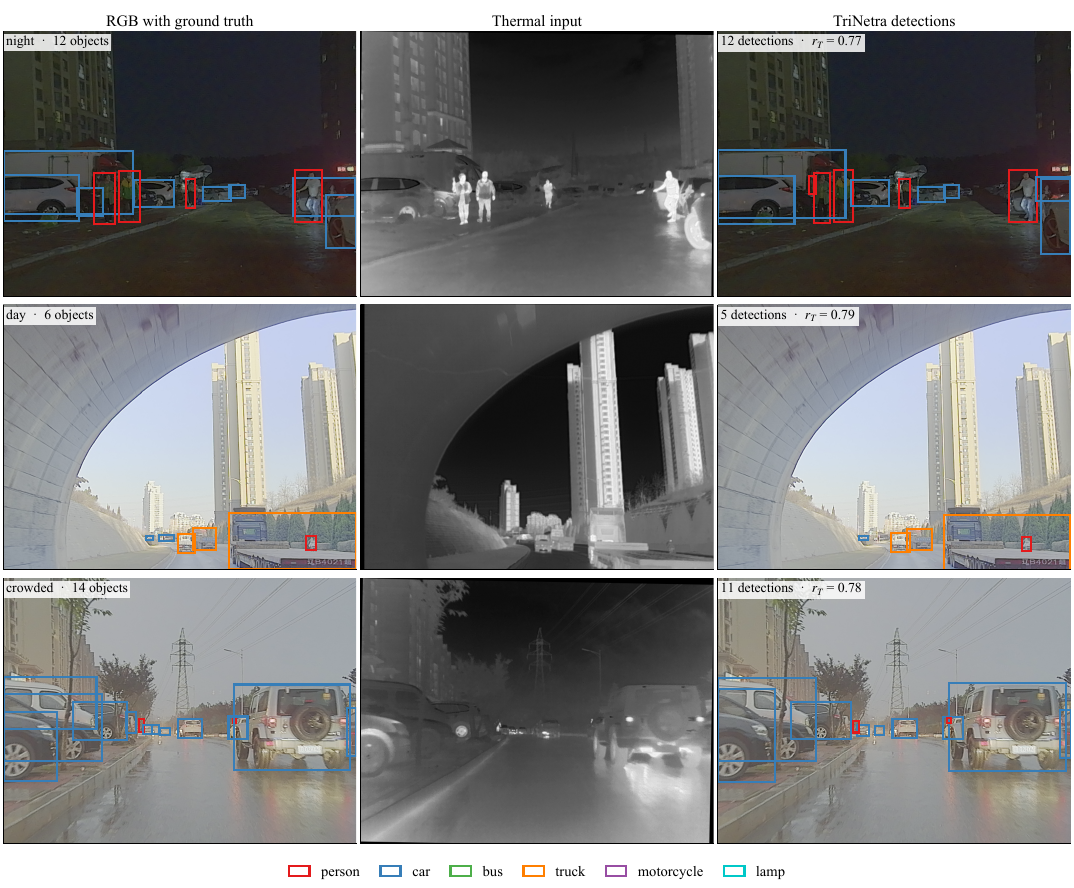}
\caption{Detections on three M3FD test pairs selected by a fixed brightness and object-count rule (not by detection quality): night, day and crowded. Left: RGB with ground truth; centre: thermal input; right: TriNetra detections at confidence 0.25.}
\label{fig:qualdet}
\end{figure*}

\begin{figure*}[!t]
\centering
\includegraphics[width=\textwidth]{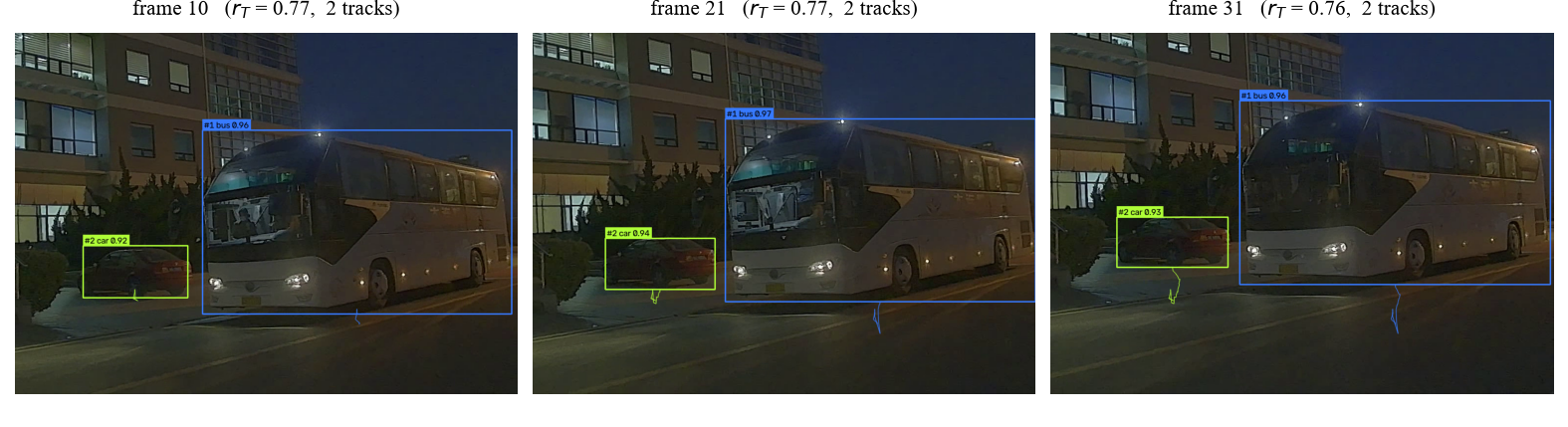}
\caption{Tracking on the longest night clip of the 14 (42 frames, mean RGB brightness 47; RGB shown). Frames at one quarter, one half and three quarters of the clip are shown. Boxes show track identity and class, and the lines show the trajectory over the preceding frames.}
\label{fig:trkqual}
\end{figure*}

Fig.~\ref{fig:qualdet} shows detections on three test pairs chosen by a fixed rule before any detection was inspected (darkest frame, brightest frame, and a mid-brightness crowded frame, each with at least six objects). In the night scene the thermal image shows the pedestrians as bright silhouettes against a dark background; all annotated pedestrians are detected and 10 of 12 objects are matched at IoU 0.5. The two misses are cars in a cluster of overlapping parked vehicles, where the predicted box differs from the annotation, and two low-score extra detections appear. In the day scene 5 of 6 objects are matched (the miss is a distant car of 774\,px$^2$), and in the crowded scene 12 of 14 (the two misses are small cars of 588 and 1,711\,px$^2$). All misses in these three frames are cars although car is the strongest class in Table~\ref{tab:detection}, so three frames should not be read as a statistical statement. Fig.~\ref{fig:trkqual} shows tracking on the longest night clip (42 frames): the bus (\#1) and the car (\#2) keep their identities in all three frames shown while the frame TRC stays between 0.76 and 0.77; the clip produced three tracks in total.

\subsection{Discussion and Limitations}
The evidence supports three claims: the TRC responds to thermal saturation, contrast loss and blur (Table~\ref{tab:sweep}); a gate driven by it makes the detector more robust to such corruption than a fixed gate with identical weights (Table~\ref{tab:gate}); and the complete pipeline runs end to end on the clips. It does not support a claim that the TRC improves tracking, that it detects misregistration, or that TriNetra is competitive in efficiency. The limitations of the present study are the following.
\begin{itemize}
\item One run, one seed, and a test split of 630 pairs in which some classes have under 110 boxes. The confidence intervals cover the choice of test images only, not run-to-run variation of training. They are about $\pm3.5$ points wide for the overall mAP@50 and up to $\pm12$ points for the rarest classes, so small differences, such as 70.0 versus 69.8, are not meaningful.
\item The gate modes are compared on one trained set of weights, the corruption is synthetic, and the network never saw corrupted data during training. Training separate single-modality and fixed-fusion models is left to future work.
\item Hand-set TRC constants ($K$, $\lambda$, equal weights). The leave-one-cue-out analysis shows that two cues compress the dynamic range on pre-registered data, and the score is blind to misalignment unless an alignment error is estimated.
\item Tracking evidence is partly simulated and small. The synthetic detector is coupled to the TRC by assumption, and on the M3FD clips that coupling is only partly confirmed. The 14 clips have no labels, are short (18 to 42 frames), are assembled from still images, and 70\% of their frames were in the detector's training split; the stress test has 28 runs on 14 clips.
\item Speed. The model needs 139.8\,ms per frame (7.2 FPS) on the laptop GPU and 408.4\,ms on the CPU through ONNX Runtime, which is not real time.
\item Training irregularities (person-only supervision in the first 35 epochs, repeated resumes, AMP, no mosaic) mean that the reported accuracy is likely below what a clean schedule would give.
\end{itemize}

\section{Conclusion and Future Work}
TriNetra couples an explicit, label-free thermal reliability score with a dual-branch detector and a tracker. The detector reaches 70.0\% mAP@50 and 47.8\% mAP@50:95 on the M3FD test split while training 17.0\,M of 42.9\,M parameters. The score decreases with thermal saturation and blur, and using it as a gate keeps 43.6\% mAP@50 under thermal corruption where a fixed gate keeps 38.4\%. The same score gave no gain in tracking and slightly lowered identity consistency under thermal saturation, and it does not detect RGB-thermal misalignment. Future work will train separate single-modality and fixed-fusion baselines and repeat runs with several seeds, train with corrupted thermal frames so that the network learns to use the gate, estimate the alignment error so that misregistration lowers the score, re-weight the cues, annotate a tracking set to measure identity accuracy on the clips, and add the planned threat-reasoning stage.

\section*{Code and Data Availability}
The experiments use the public M3FD dataset \cite{m3fd}. The code, configuration files, evaluation scripts and the trained detector (release v1.0, SHA-256 \texttt{e7fac200784a5518c5b859\allowbreak effdadac8764484cc1bd70\allowbreak 823ee9489f4d9a6bb124}) are available at \url{https://github.com/vedant-n-sonawane/Tri-Netra} for research and educational use. The M3FD clips used for the tracking tests can be rebuilt with \texttt{utils/make\_m3fd\_clips.py}, which also writes the first and last file of every clip.

\section*{Acknowledgment}
The authors thank the creators of the M3FD dataset for making it publicly available.

\balance


\begin{thebibliography}{99}

\bibitem{songSurvey}
K.~Song, Y.~Zhao, L.~Huang, Y.~Yan, and Q.~Meng, ``{RGB-T} image analysis technology and application: A survey,'' \textit{Eng. Appl. Artif. Intell.}, vol.~120, Art. no.~105919, 2023, doi: 10.1016/j.engappai.2023.105919.

\bibitem{deepFusion}
F.~Zhao, C.~Zhang, and B.~Geng, ``Deep multimodal data fusion,'' \textit{ACM Comput. Surv.}, vol.~56, no.~9, Art. no.~216, 2024, doi: 10.1145/3649447.

\bibitem{denet}
Q.~Wang, Y.~Sun, Y.~Chi, and T.~Shen, ``{RGB-T} object detection with failure scenarios,'' \textit{IEEE J. Sel. Topics Appl. Earth Observ. Remote Sens.}, vol.~18, pp.~3000--3010, 2025, doi: 10.1109/JSTARS.2024.3523408.

\bibitem{qdfm}
H.~Yu, R.~Luo, X.~Wang, D.~Wang, Q.~Li, and S.~Qiao, ``Robust {RGB-T} object detection via quality-aware decision fusion based on {Dempster--Shafer} theory,'' \textit{IEEE Sensors J.}, vol.~25, no.~13, pp.~26051--26064, 2025, doi: 10.1109/JSEN.2025.3566381.

\bibitem{tfdet}
X.~Zhang, X.~Zhang, J.~Wang, J.~Ying, Z.~Sheng, H.~Yu, C.~Li, and H.-L.~Shen, ``{TFDet}: Target-aware fusion for {RGB-T} pedestrian detection,'' \textit{IEEE Trans. Neural Netw. Learn. Syst.}, vol.~36, no.~7, 2025, doi: 10.1109/TNNLS.2024.3443455.

\bibitem{gemyolo}
L.~Wang, Z.~Bao, and D.~Lu, ``{GEM-YOLO}: A lightweight and real-time {RGBT} object detector with gated multimodal fusion,'' \textit{Sensors}, vol.~26, no.~7, Art. no.~2035, 2026, doi: 10.3390/s26072035.

\bibitem{eef}
T.~Hao, J.~Yang, S.~Zhang, and S.~Wu, ``{EEF}: Energy score-guided feature enhancement fusion method for {RGB} and thermal infrared images object detection,'' \textit{Signal Process.}, vol.~239, Art. no.~110231, 2026, doi: 10.1016/j.sigpro.2025.110231.

\bibitem{yolov8}
G.~Jocher, A.~Chaurasia, and J.~Qiu, ``Ultralytics {YOLOv8},'' 2023. [Online]. Available: https://github.com/ultralytics/ultralytics

\bibitem{bytetrack}
Y.~Zhang \textit{et al.}, ``{ByteTrack}: Multi-object tracking by associating every detection box,'' in \textit{Proc. Eur. Conf. Comput. Vis. (ECCV)}, 2022, pp.~1--21.

\bibitem{yolo}
J.~Redmon, S.~Divvala, R.~Girshick, and A.~Farhadi, ``You only look once: Unified, real-time object detection,'' in \textit{Proc. IEEE Conf. Comput. Vis. Pattern Recognit. (CVPR)}, 2016, pp.~779--788.

\bibitem{sort}
A.~Bewley, Z.~Ge, L.~Ott, F.~Ramos, and B.~Upcroft, ``Simple online and realtime tracking,'' in \textit{Proc. IEEE Int. Conf. Image Process. (ICIP)}, 2016, pp.~3464--3468.

\bibitem{kalman}
R.~E. Kalman, ``A new approach to linear filtering and prediction problems,'' \textit{J. Basic Eng.}, vol.~82, no.~1, pp.~35--45, 1960.

\bibitem{kuhn}
H.~W. Kuhn, ``The Hungarian method for the assignment problem,'' \textit{Naval Res. Logist. Quart.}, vol.~2, nos.~1--2, pp.~83--97, 1955.

\bibitem{deepsort}
N.~Wojke, A.~Bewley, and D.~Paulus, ``Simple online and realtime tracking with a deep association metric,'' in \textit{Proc. IEEE Int. Conf. Image Process. (ICIP)}, 2017, pp.~3645--3649.

\bibitem{hota}
J.~Luiten \textit{et al.}, ``{HOTA}: A higher order metric for evaluating multi-object tracking,'' \textit{Int. J. Comput. Vis.}, vol.~129, no.~2, pp.~548--578, 2021.

\bibitem{clearmot}
K.~Bernardin and R.~Stiefelhagen, ``Evaluating multiple object tracking performance: The {CLEAR MOT} metrics,'' \textit{EURASIP J. Image Video Process.}, vol.~2008, Art. no.~246309, 2008.

\bibitem{idf1}
E.~Ristani, F.~Solera, R.~Zou, R.~Cucchiara, and C.~Tomasi, ``Performance measures and a data set for multi-target, multi-camera tracking,'' in \textit{Proc. Eur. Conf. Comput. Vis. Workshops (ECCVW)}, 2016, pp.~17--35.

\bibitem{m3fd}
J.~Liu, X.~Fan, Z.~Huang, G.~Wu, R.~Liu, W.~Zhong, and Z.~Luo, ``Target-aware dual adversarial learning and a multi-scenario multi-modality benchmark to fuse infrared and visible for object detection,'' in \textit{Proc. IEEE/CVF Conf. Comput. Vis. Pattern Recognit. (CVPR)}, 2022, pp.~5802--5811.

\bibitem{bilateral}
C.~Tomasi and R.~Manduchi, ``Bilateral filtering for gray and color images,'' in \textit{Proc. IEEE Int. Conf. Comput. Vis. (ICCV)}, 1998, pp.~839--846.

\bibitem{clahe}
K.~Zuiderveld, ``Contrast limited adaptive histogram equalization,'' in \textit{Graphics Gems IV}, P.~S. Heckbert, Ed. San Diego, CA, USA: Academic, 1994, pp.~474--485.

\bibitem{shannon}
C.~E. Shannon, ``A mathematical theory of communication,'' \textit{Bell Syst. Tech. J.}, vol.~27, no.~3, pp.~379--423, 1948.

\bibitem{pech}
J.~L. Pech-Pacheco, G.~Crist\'obal, J.~Chamorro-Mart\'inez, and J.~Fern\'andez-Valdivia, ``Diatom autofocusing in brightfield microscopy: A comparative study,'' in \textit{Proc. 15th Int. Conf. Pattern Recognit. (ICPR)}, 2000, vol.~3, pp.~314--317.

\bibitem{viola}
P.~Viola and W.~M. Wells, III, ``Alignment by maximization of mutual information,'' \textit{Int. J. Comput. Vis.}, vol.~24, no.~2, pp.~137--154, 1997.

\bibitem{pytorch}
A.~Paszke \textit{et al.}, ``{PyTorch}: An imperative style, high-performance deep learning library,'' in \textit{Proc. Adv. Neural Inf. Process. Syst. (NeurIPS)}, 2019, pp.~8024--8035.

\bibitem{sgdr}
I.~Loshchilov and F.~Hutter, ``{SGDR}: Stochastic gradient descent with warm restarts,'' in \textit{Proc. Int. Conf. Learn. Represent. (ICLR)}, 2017.

\bibitem{coco}
T.-Y. Lin \textit{et al.}, ``Microsoft {COCO}: Common objects in context,'' in \textit{Proc. Eur. Conf. Comput. Vis. (ECCV)}, 2014, pp.~740--755.

\bibitem{efron}
B.~Efron and R.~J. Tibshirani, \textit{An Introduction to the Bootstrap}. New York, NY, USA: Chapman \& Hall, 1993.

\bibitem{wilcoxon}
F.~Wilcoxon, ``Individual comparisons by ranking methods,'' \textit{Biometrics Bull.}, vol.~1, no.~6, pp.~80--83, 1945.

\end{thebibliography}
\end{document}